\documentclass{article}

\usepackage{amsmath}
\usepackage[shortlabels]{enumitem}
\usepackage{subcaption}
\usepackage{float}
\usepackage{graphicx}

\PassOptionsToPackage{numbers,compress}{natbib}
\usepackage{neurips_2025}

\usepackage[utf8]{inputenc} 
\usepackage[T1]{fontenc}    
\usepackage{hyperref}       
\usepackage{url}            
\usepackage{booktabs}       
\usepackage{amsfonts}       
\usepackage{nicefrac}       
\usepackage{microtype}      
\usepackage{xcolor}         
\usepackage{orcidlink}

\title{BULL-ODE: Bullwhip Learning with Neural ODEs and Universal Differential Equations under Stochastic Demand}

\makeatletter\@anonymousfalse\makeatother
\author{
  Nachiket N. Naik~\orcidlink{https://orcid.org/0009-0008-9462-870X}\\
  Columbia University (Prior) \\
  \texttt{nachiket.n.naik@gmail.com} \\
  \And
  Prathamesh Dinesh Joshi~\orcidlink{https://orcid.org/009-0008-5476-9962}\\
  Vizuara AI Labs \\
  \texttt{prathamesh@vizuara.com} \\
  \AND
  Raj Abhijit Dandekar \\
  Vizuara AI Labs \\
  \texttt{raj@vizuara.com} \\
  \And
  Rajat Dandekar \\
  Vizuara AI Labs \\
  \texttt{rajatdandekar@vizuara.com} \\
  \And
  Sreedath Panat \\
  Vizuara AI Labs \\
  \texttt{sreedath@vizuara.com} \\
}

\begin{document}
\maketitle

\begin{abstract}
We study learning of continuous‑time inventory dynamics under stochastic demand and quantify when structure helps or hurts forecasting of the bullwhip effect. \emph{BULL‑ODE} compares a fully learned Neural ODE (NODE) that models the entire right‑hand side against a physics‑informed Universal Differential Equation (UDE) that preserves conservation and order‑up‑to structure while learning a small residual policy term. Classical supply chain models explain the bullwhip through control/forecasting choices and information sharing, while recent physics-informed and neural differential equation methods blend domain constraints with learned components. It is unclear whether structural bias helps or hinders forecasting under different demand regimes. We address this by using a single-echelon testbed with three demand regimes - AR(1) (autocorrelated), i.i.d. Gaussian, and heavy‑tailed lognormal. Training is done on varying fractions of each trajectory, followed by evaluation of multi‑step forecasts for inventory $I$, order rate $O$, and demand $D$. Across the structured regimes, UDE consistently generalizes better: with 90\% of the training horizon, inventory RMSE drops from 4.92 (NODE) to 0.26 (UDE) under AR(1) and from 5.96 to 0.95 under Gaussian demand. Under heavy‑tailed lognormal shocks, the flexibility of NODE is better. These trends persist as training data shrinks, with NODE exhibiting phase drift in extrapolation while UDE remains stable but underreacts to rare spikes. Our results provide concrete guidance: enforce structure when noise is light‑tailed or temporally correlated; relax structure when extreme events dominate. Beyond inventory control, the results offer guidance for hybrid modeling in scientific and engineering systems: enforce known structure when conservation laws and modest noise dominate, and relax structure to capture extremes in settings where rare events drive dynamics.
\end{abstract}

\section{Introduction}

Small fluctuations in retail demand can be dramatically amplified as orders propagate upstream through distributors and manufacturers. This "bullwhip effect" has been observed since early industrial dynamics work and remains a central concern because it increases costs, lengthens lead times, and degrades service \cite{Forrester1961, Lee1997}. Laboratory evidence from the Beer Game shows that human decision makers systematically misperceive delays and feedback, further fueling oscillations and variance amplification \cite{Sterman1989}. Subsequent experiments verify these behavioral mechanisms and quantify the benefit of inventory information sharing \cite{CrosonDonohue2006}.

Most accounts trace the bullwhip effect to four main causes: demand signal processing, order batching, price variations, and rationing \cite{Lee1997}. Research shows that forecasting methods, inventory rules, and lead times interact to amplify variability along the chain. Studies have also quantified the benefits of upstream–downstream information sharing in two-level supply chains \cite{LeeSoTang2000}. Even in a simple two-stage setting, longer replenishment lead times or the use of naive forecasting rules can increase the order-to-demand variance ratio, despite stable underlying demand \cite{Chen2000MS, Chen2000NRL}.

Control-theoretic analyses reveal that common order-up-to policies are predisposed to bullwhip under broad conditions, and they provide closed-form expressions linking amplification to policy parameters and lead times \cite{DisneyTowill2003Omega, Dejonckheere2003, Dejonckheere2004}. Extensions quantify the role of multi-tier structure and MMSE forecasting, showing that accumulated lead times—rather than the sheer number of stages—govern amplification \cite{HosodaDisney2006}. Forecasting-method choices (e.g., moving average vs. exponential smoothing) shape both the magnitude and sensitivity of bullwhip \cite{Zhang2004, Chen2000NRL}.

Neural differential equation models have emerged as a powerful tool to learn system dynamics directly from the data while retaining some benefits of continuous-time formulations. Neural Ordinary Differential Equations (NODEs) treat the time-evolution of system state as a learned function, $\frac{d\mathbf{x}}{dt} = f_{\theta}(\mathbf{x}, t)$, where $f_{\theta}$ is a neural network \cite{Chen2018}. NODEs have shown promise in modeling complex time-series and physical processes without requiring an explicit mechanistic model. However, a pure NODE may overfit noise or violate physical constraints when data is scarce or when the system has a known structure.

To address this issue, the Universal Differential Equations (UDE) framework integrates neural networks into known differential equation structures. This combines domain knowledge with learning \cite{Rackauckas2020}. In a UDE, we model the complete ODE using a neural network to capture dynamics that are unknown or too complex to model analytically. This approach is a part of the broader field of scientific machine learning. Related physics-informed formulations include PINNs \cite{Raissi2019}. It aims to enforce physical consistency while still fitting complex patterns from data. Our work applies the NODE and UDE approaches to a classic supply chain inventory model to compare their effectiveness in capturing the bullwhip effect under various stochastic demand scenarios.

In this paper, we present a comparison of a learning-based NODE model and a physics-informed UDE model on an inventory control system. We focus on how each performs under different types of demand uncertainty and with different amounts of training data:

\begin{itemize}
    \item \textbf{Testbed:} A single-level supply chain model that shows the bullwhip effect - a continuous system for managing inventory and orders with feedback control.

    \item \textbf{Stochastic demand processes:} different demand patterns of are considered:
        \begin{enumerate}
        \item an autocorrelated AR(1) demand (mimicking short-term demand correlation),
        \item uncorrelated Gaussian noise (baseline random fluctuations), and
        \item lognormal multiplicative noise (heavy-tailed demand shocks).
        \end{enumerate}

    \item \textbf{Evaluation protocol:} models are trained on varying portions of the time series to test generalization. We evaluated \textit{ training fit} and \textit{forecasting accuracy} on held-out periods using RMSE for key state variables: inventory level, order rate, and demand rate.
    
\end{itemize}

\section{Methodology}
\label{sec:headings}

We consider a single-echelon inventory system in continuous time, from classical industrial dynamics and control theory formulations of inventory control and order-up-to policies \cite{Forrester1961, DisneyTowill2003Omega, Dejonckheere2003, HosodaDisney2006}. The state variables are inventory $I(t)$, ordering rate $O(t)$ (orders placed by the retailer to replenish inventory), and customer demand $D(t)$. The base dynamics are:

\begin{align}
\frac{dI(t)}{dt} &= O(t) - D(t), \label{eq:inventory_balance}\\
\frac{dO(t)}{dt} &= \frac{1}{\tau}\Big[D(t) + \alpha\big(I_{\text{target}} - I(t)\big) - O(t)\Big], \label{eq:ordering_policy}
\end{align}

where $\tau$ is an inventory adjustment time constant and $\alpha$ is a stock correction coefficient that determines how aggressively the company corrects inventory deviations.

Equation~\eqref{eq:inventory_balance} is the standard inventory conservation law (inflow minus outflow) \cite{Forrester1961}. The term $D(t) + \alpha(I_{\text{target}}-I)$ in Eq.(\ref{eq:ordering_policy}) represents a simple order-up-to policy: the firm orders to satisfy current demand and to slowly eliminate the gap between the actual inventory $I(t)$ and a target level $I_{\text{target}}$. This demand plus inventory correction structure is widely analyzed in bullwhip studies \cite{DisneyTowill2003Omega, Dejonckheere2003, Dejonckheere2004, HosodaDisney2006}. This linear feedback can generate oscillations and variance amplification (bullwhip) when $\alpha$ or $\tau$ are poorly tuned or when lead times/forecasting interact adversely \cite{DisneyTowill2003Omega, Dejonckheere2003, Chen2000MS, Chen2000NRL}. We set $I_{\text{target}}$ to a nominal level (e.g., the initial inventory) so that the system is stationary in the absence of demand fluctuations.

The pair \eqref{eq:inventory_balance}–\eqref{eq:ordering_policy} corresponds to the continuous-time order-up-to family used to study bullwhip and stability in production–inventory systems; see \cite{Forrester1961, DisneyTowill2003Omega, Dejonckheere2003, Dejonckheere2004, HosodaDisney2006} for derivations and closed-form variance/amplification results.

\subsection{Stochastic demand processes}
In our experiments, we simulate the customer demand $D(t)$ using the following processes to stress-test the models:

\begin{itemize}
    \item \textbf{AR(1) Correlated Demand:} $D(t)$ follows a first-order autoregressive process, where demand has memory:
    \[
        D_{n+1} = \mu + \phi (D_n - \mu) + \varepsilon_n.
    \]
    We choose $\phi = 0.6$ (moderate positive autocorrelation) and Gaussian shocks $\varepsilon_n$ such that $\mathrm{Var}(\varepsilon) = \sigma^2$ (e.g., $\sigma = 3$ in our setting). This produces temporally correlated demand swings. In continuous-time simulation, we approximate this by an Ornstein–Uhlenbeck process for $D(t)$ with mean $\mu$ and relaxation rate related to $\phi$. Such models are naturally treated within a state-space framework, where estimation methods are classical \cite{DurbinKoopman2012}.

    \item \textbf{Gaussian white noise demand:} $D(t)$ fluctuates as an i.i.d.\ random process with no memory. At each (discrete) time step, demand is drawn from a Gaussian distribution $N(\mu, \sigma^2)$. We use $\sigma = 5$ for a high-variance case. This scenario represents \emph{erratic but light-tailed demand}. For training, we treat the realized demand sequence as given, and evaluation uses the same draws for all models. Forecasting benchmarks under such conditions are well studied \cite{MakridakisHibon2000, HyndmanAthanasopoulos2018}, and the Box–Jenkins ARIMA framework remains a canonical baseline for light-tailed demand \cite{BoxJenkins2015}.

    \item \textbf{Lognormal heavy-tailed demand:} To simulate occasional extreme demand spikes, we use a lognormal distribution for demand increments. Specifically, we generate multiplicative shocks such that
    \[
        \ln(D / \mu) \sim N(0, \tilde{\sigma}^2)
    \]
    with $\tilde{\sigma} = 2$ in our tests, so that $D(t)$ is usually near $\mu$ but with rare large bursts. This yields a heavy-tailed demand pattern \cite{Clauset2009, Embrechts1997}. Such dynamics are challenging for models assuming smooth or Gaussian noise. Jump-diffusion and SDE formulations are classical approaches for capturing discontinuities \cite{Merton1976, KloedenPlaten1992}, while neural SDE variants provide differentiable, learnable counterparts \cite{TzenRaginsky2019}.
    
\end{itemize}

For each noise type, we synthesize a time series of length $T$ by numerically solving the inventory ODEs. We use a fixed stepsize integrator (e.g., 0.2 time-unit step with a Tsit5 solver), specifically the Tsit5 Runge–Kutta pair for time stepping \cite{Tsitouras2011}, and inject the stochastic demand behavior at each step. The result is a set of trajectories $I(t), O(t), D(t)$ that exhibit the bullwhip effect to varying degrees depending on the demand noise characteristics. Figure~1 illustrates sample trajectories of inventory and orders under these demand processes, showing how autocorrelation or heavy tails in $D(t)$ lead to different inventory oscillation patterns.

\paragraph{Hyperparameter tuning and selected settings.}
We tuned the stochastic process and model hyperparameters on a held-out validation slice at the end of each training window, selecting values that minimized validation RMSE and produced stable trajectories across regimes. The search covered small grids around standard choices (e.g., $\phi\in\{0.3,0.6,0.8\}$, $\sigma\in\{2,3,5,7\}$, $\tilde\sigma\in\{1,2,2.5\}$, $\tau\in\{3,5,8\}$, $\alpha\in\{0.4,0.6,0.8,1.0\}$, and network widths $\in\{16,32,64\}$). The selected settings used in all reported results are summarized in Table~\ref{tab:hparams}.

\begin{table}[H]
\centering
\small
\caption{Hyperparameters selected after tuning.}
\label{tab:hparams}
\begin{tabular}{ll}
\toprule
\textbf{Component} & \textbf{Selected value} \\
\midrule
AR(1) coefficient $\phi$ & $0.6$ \\
AR(1) shock std.\ $\sigma$ & $3$ \\
Gaussian demand std.\ $\sigma$ & $5$ \\
Lognormal multiplicative std.\ $\tilde\sigma$ & $2$ \\
Inventory adjustment time $\tau$ & $5$ \\
Stock correction $\alpha$ & $0.8$ \\
NODE architecture & 2 hidden layers $\times$ 64 units, $leakyrelu$ \\
UDE correction $g_{\phi}$ & 2 hidden layers $\times$ 16 units, $\tanh$ \\
Time step $\Delta t$ & $0.2$ \\
Initial/target inventory $I_{\text{target}}$ & $100$ \\
Nominal rate $\mu$ & $10$ \\
\bottomrule
\end{tabular}
\end{table}

\subsection{Neural ODE (NODE) model}

The NODE approach learns the entire right-hand side of the dynamics as a black-box neural network. We define a trainable function $f_{\theta}:\mathbb{R}^3 \to \mathbb{R}^3$ such that $(\dot I,,\dot O,,\dot D) = f_{\theta}(I, O, D)$. We are using a feedforward neural network (with $2$ hidden layers and tens of neurons per layer) to represent $f_{\theta}$. This network models all interactions and feedback among inventory, orders, and demand directly from data, without being given Eq.(\ref{eq:inventory_balance}) or (\ref{eq:ordering_policy}) explicitly.

During training, the NODE’s parameters $\theta$ are optimized so that the solution of the neural ODE matches the observed trajectory over the training period. The ODE is integrated with an adaptive solver, and gradients are computed using the adjoint sensitivity method \cite{Chen2018}. We adopt a two-phase schedule: (i) \emph{Adam} to explore the nonconvex landscape with robust, scale-adapted steps \cite{KingmaBa2015}, followed by (ii) \emph{BFGS} with line search to exploit local curvature and drive the MSE to a low training loss on this smooth ODE-constrained objective \cite{Fletcher1970, Shanno1970, BottouCurtisNocedal2018}. This Adam–BFGS hand-off is common in scientific machine learning and neural-ODE pipelines. It improves convergence speed and final fit \cite{Rackauckas2020}, and closely parallels established practice in system identification \cite{Ljung1999}.

\subsection{Universal differential equation (UDE) model}

The UDE approach uses knowledge of the system’s structure by starting from Eqs.(\ref{eq:inventory_balance})–(\ref{eq:ordering_policy}) and only learning the parts that are unknown or too complex. In our UDE design, we retain the inventory balance equation exactly (Eq. 1) and the general form of the ordering policy (Eq.2), and introduce a neural network term to model the order rate dynamics. Specifically, we modify Eq. (\ref{eq:ordering_policy}) to:

\begin{equation}\label{eq:ude_order}
\frac{dO(t)}{dt} = \frac{1}{\tau}\Big[ D(t) + \alpha (I_{\text{target}} - I(t)) - O(t)\Big] + g_{\phi}(I(t), O(t), D(t))
\end{equation}

where $g_{\phi}$ is a small neural network (with parameters $\phi$) that learns changes in the ordering decision rule. The function $g_{\phi}$ (e.g., a 2-layer perceptron with 16 hidden units) takes the current state $(I, O ,D)$ as input and outputs an adjustment to $dO/dt$. This can capture unmodeled nonlinear effects or an adaptive policy that the simple linear law misses. 

Importantly, $g_{\phi}$ is initialized with small weights so that the model output doesn't explode. The UDE assumes we know how $D(t)$ evolves in expectation. This reflects a common scenario in supply chains: demand is considered an input rather than a state to predict. The UDE focuses on learning the inventory/order dynamics (i.e., the replenishment policy). We train the UDE by integrating the augmented system (Eq. \ref{eq:inventory_balance} and Eq. \ref{eq:ude_order}) over the training horizon and minimizing MSE between $(I, O)$ predicted vs. true. Training uses the same strategy as NODE (adjoint gradients and hybrid optimization). The result is a model that respects known relationships and generalizes better, while still using the neural term $g_{\phi}$ to fit complex behaviors that Eq.(\ref{eq:ordering_policy}) alone cannot capture.

\subsection{Experimental setup and evaluation protocol}

We generate simulated data for each of the demand scenarios (AR(1), Gaussian, lognormal) using the process described above. For consistency, all simulations use the same initial conditions (e.g. $I(0)=I_{\text{target}}=100$ units, $O(0)=D(0)=\mu=10$ units/time) and model parameters ($\tau$ and $\alpha$ chosen to give a stable but responsive system; e.g. $\tau = 5$ time units, $\alpha = 0.8$). Each simulation produces time series for $I(t)$, $O(t)$, and $D(t)$ for $t \in [0,T]$.

We partition each time series into a training segment $[0, T_{\text{train}}]$ and a testing (forecast) segment $(T_{\text{train}}, T]$. We consider multiple train-test splits: specifically, $T_{\text{train}} = p,T$ with $p \in \small({0.5, 0.6, 0.7, 0.8, 0.9}\small)$. This allows us to analyze performance both in a data-rich regime ($p=0.9$) and a data-scarce regime ($p=0.5$), and intermediates. For each training fraction $p$, we train separate NODE and UDE models on the same training subset and then evaluate them on the same forecasting task. To ensure a fair comparison, both models are trained using identical optimization settings and stopping criteria.

\subsubsection{Evaluation metrics}

During training, we record the final training loss (MSE) for NODE vs UDE to gauge how well each fits the training data. However, our primary evaluation metric is the forecast RMSE in the test period for each state variable: inventory $I$, order rate $O$, and demand $D$. The forecast RMSE for a variable $X$ is defined as $\sqrt{\frac{1}{N}\sum_{t=T_{\text{train}}}^{T} \big(\hat{X}(t) - X_{\text{true}}(t)\big)^2}$, where $\hat{X}(t)$ is the prediction of the model and $X_{\text{true}}(t)$ is the simulation of the ground truth. $\hat{X}(t)$ is obtained by integrating the learned ODE forward from $T_{\text{train}}$ without seeing true future values.

This essentially measures how far off the forecasts are, penalizing larger errors. We compute RMSE for $I$, $O$, and $D$ separately to see which aspects of system behavior are captured or missed by each modeling approach.

\begin{figure}[htbp]
  \centering
  \begin{subfigure}[t]{0.4\textwidth}
    \centering
    \includegraphics[width=\linewidth]{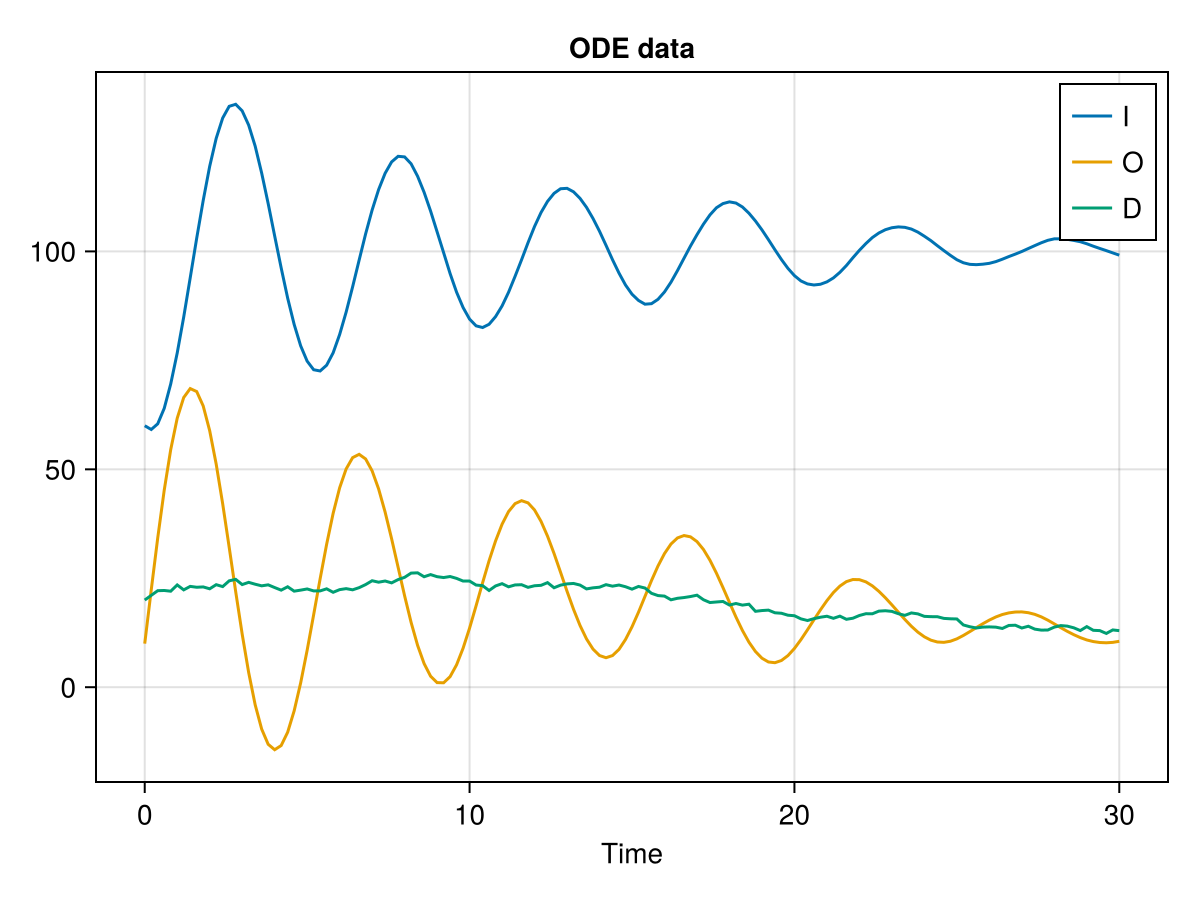}
    \caption{Autocorrelated noise}
    \label{fig:ode_autocorr}
  \end{subfigure}\hfill
  \begin{subfigure}[t]{0.4\textwidth}
    \centering
    \includegraphics[width=\linewidth]{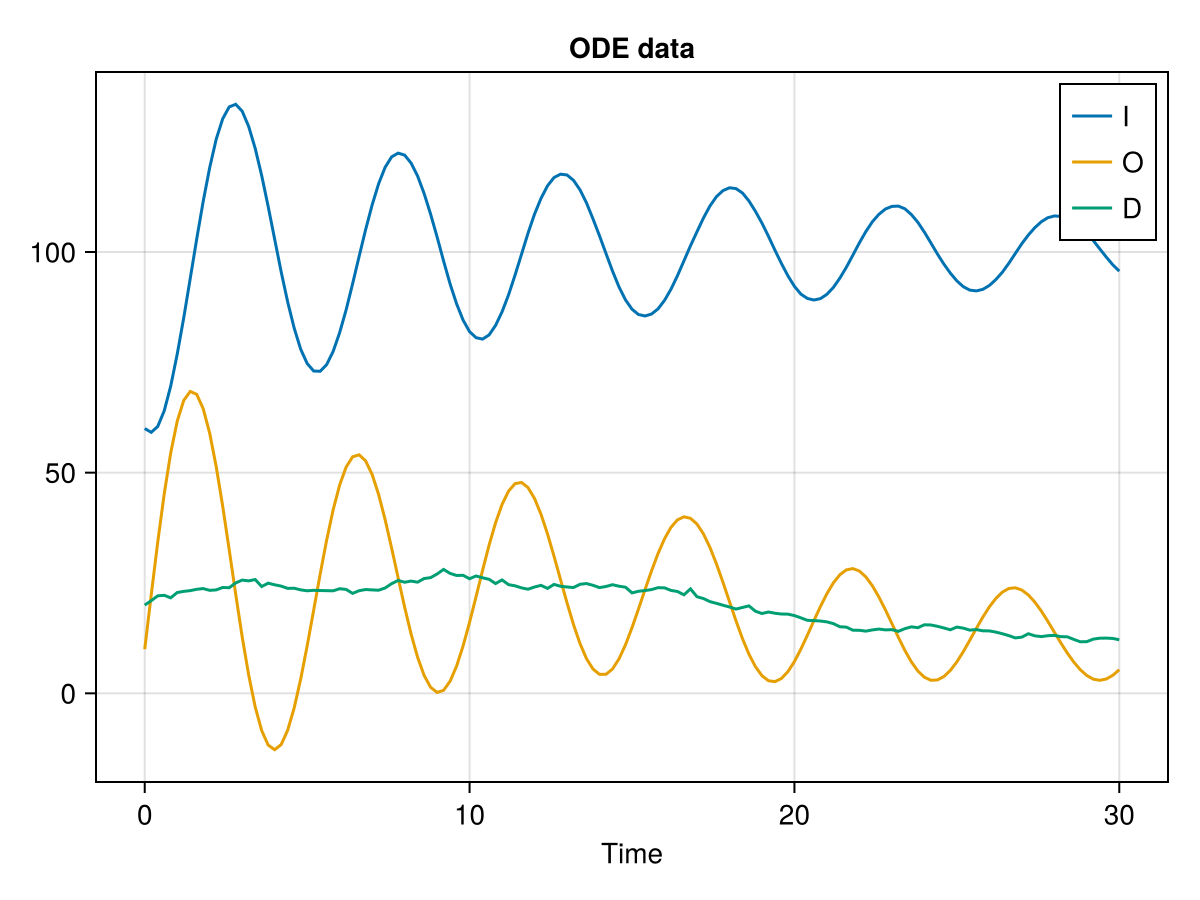}
    \caption{Gaussian noise}
    \label{fig:ode_gaussian}
  \end{subfigure}\hfill
  \begin{subfigure}[t]{0.4\textwidth}
    \centering
    \includegraphics[width=\linewidth]{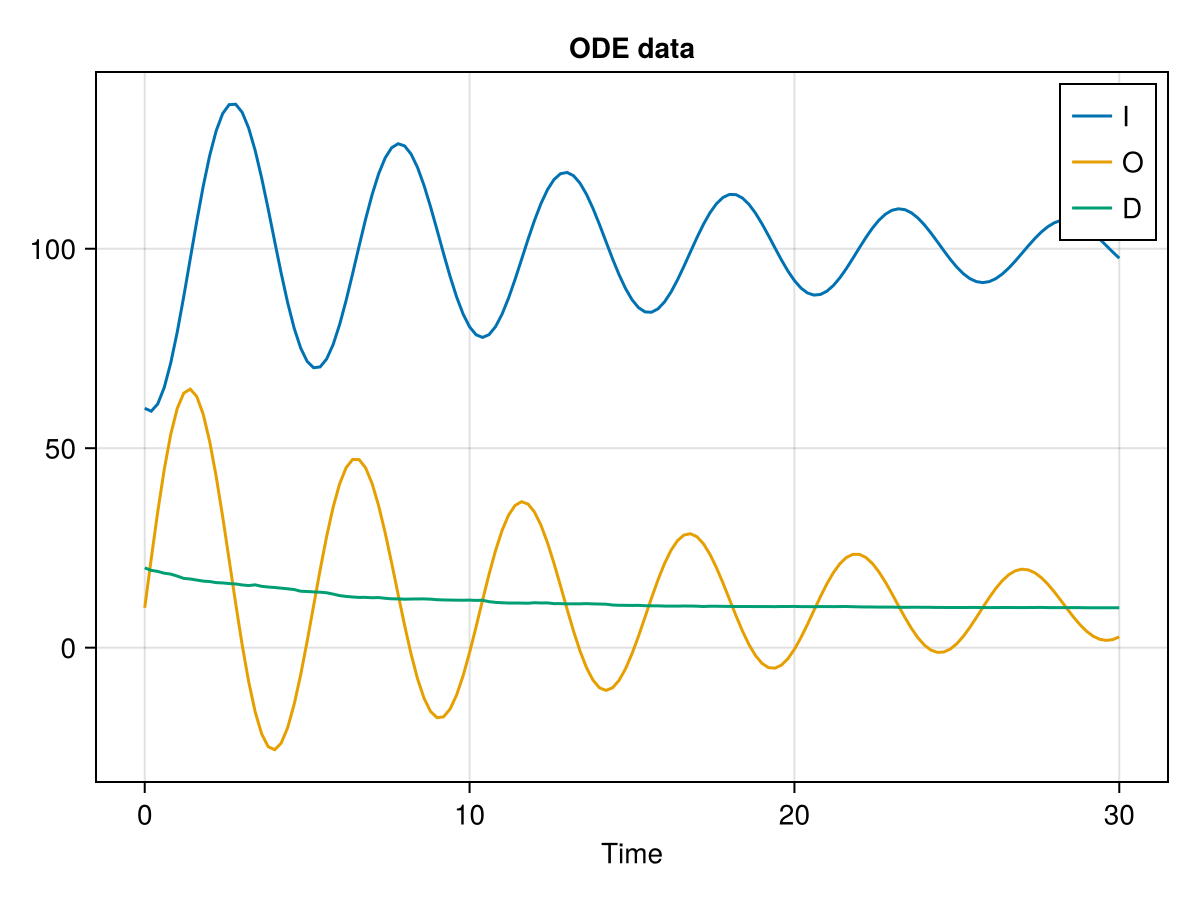}
    \caption{Lognormal noise}
    \label{fig:ode_lognormal}
  \end{subfigure}
  \caption{Simulated ODE data under three demand scenarios.}
  \label{fig:ode_scenarios}
\end{figure}

\section{Results}
\label{sec:results}

We compare a fully learned Neural ODE against a physics-informed Universal Differential Equation. We evaluate forecasting RMSE in the held-out horizon for three demand regimes - AR(1) (autocorrelated), Gaussian (i.i.d.), and Lognormal (heavy-tailed). For two representative train/test splits (90\% vs.\ 60\% of the horizon used for training).

\begin{table}[H]
\centering
\small
\caption{Forecast RMSE on the test horizon. Each entry reports RMSE for inventory $I$, order rate $O$, and demand $D$ at two train fractions.}
\vspace{0.5em}
\label{tab:rmse-summary}
\begin{tabular}{ll|ccc|ccc}
\toprule
Demand Process & Model & RMSE$_I$ & RMSE$_O$ & RMSE$_D$ & RMSE$_I$ & RMSE$_O$ & RMSE$_D$ \\
 &  & \multicolumn{3}{c|}{Train 90\% (more data)} & \multicolumn{3}{c}{Train 60\% (less data)} \\
\midrule
AR(1)      & NODE & 4.92 & 6.48 & 5.38 & 11.25 & 4.45 & 4.09 \\
AR(1)      & UDE  & 0.26 & 2.11 & 2.32 &  5.09 & 6.72 & 1.59 \\
Gaussian   & NODE & 5.96 & 7.45 & 3.61 &  2.43 & 10.58 & 8.38 \\
Gaussian   & UDE  & 0.95 & 4.08 & 3.10 &  3.77 &  5.29 & 2.16 \\
Lognormal  & NODE & 1.98 & 1.42 & 1.19 &  6.62 &  2.71 & 1.23 \\
Lognormal  & UDE  & 3.04 & 8.04 & 5.61 &  3.64 &  8.03 & 6.05 \\
\bottomrule
\end{tabular}
\end{table}

\paragraph{High-level trends.} Table~\ref{tab:rmse-summary} shows a consistent pattern:

\begin{enumerate}[(i)]
\item AR(1) (autocorrelated demand): UDE substantially outperforms NODE. With 90\% training data, UDE reduces inventory RMSE from $4.92$ (NODE) to $0.26$ ($\approx95\%$ lower), and similarly improves $O$ and $D$. Even at 60\% training, UDE maintains strong $I$ and $D$ accuracy, although $O$ gets noisier.

\item Gaussian (i.i.d.): UDE again outperforms NODE; at 90\% training, inventory RMSE improves from $5.96$ for NODE to $0.95$ ($\approx84\%$ lower). With 60\% training, NODE’s $I$ RMSE is close to UDE ($2.43$ vs.\ $3.77$), but NODE’s $O$ and $D$ errors grow significantly.

\item Lognormal (heavy-tailed): NODE performs significantly better, especially on $O$: at 90\% training, NODE’s order RMSE is $1.42$ vs.\ UDE’s $8.04$ ($\approx82\%$ lower). The structural bias of UDE underreacts to rare demand spikes.
\end{enumerate}

\subsubsection*{Interpretation by Regime}

\textit{AR(1) and Gaussian.}
Because UDE enforces the conservation law and the stabilizing order-up-to policy, it filters stochastic fluctuations and generalizes robustly beyond the training window. NODE, which learns the full RHS, achieves very low training error but tends to fit high-frequency noise that does not recur in the test segment, leading to drift or variance amplification during extrapolation. The gap widens as the training fraction decreases.

\textit{Lognormal.}
Here, sharp multiplicative shocks dominate the dynamics. The flexible NODE adapts to those extremes and forecasts large order surges, whereas UDE’s small additive correction on \(\dot O\) cannot reproduce rare spikes well, yielding larger order and downstream inventory errors.

\vspace{-0.7\baselineskip}

\begin{figure}[htbp]
\centering
    \begin{subfigure}[t]{0.48\textwidth}
    \centering
    \includegraphics[width=\linewidth]{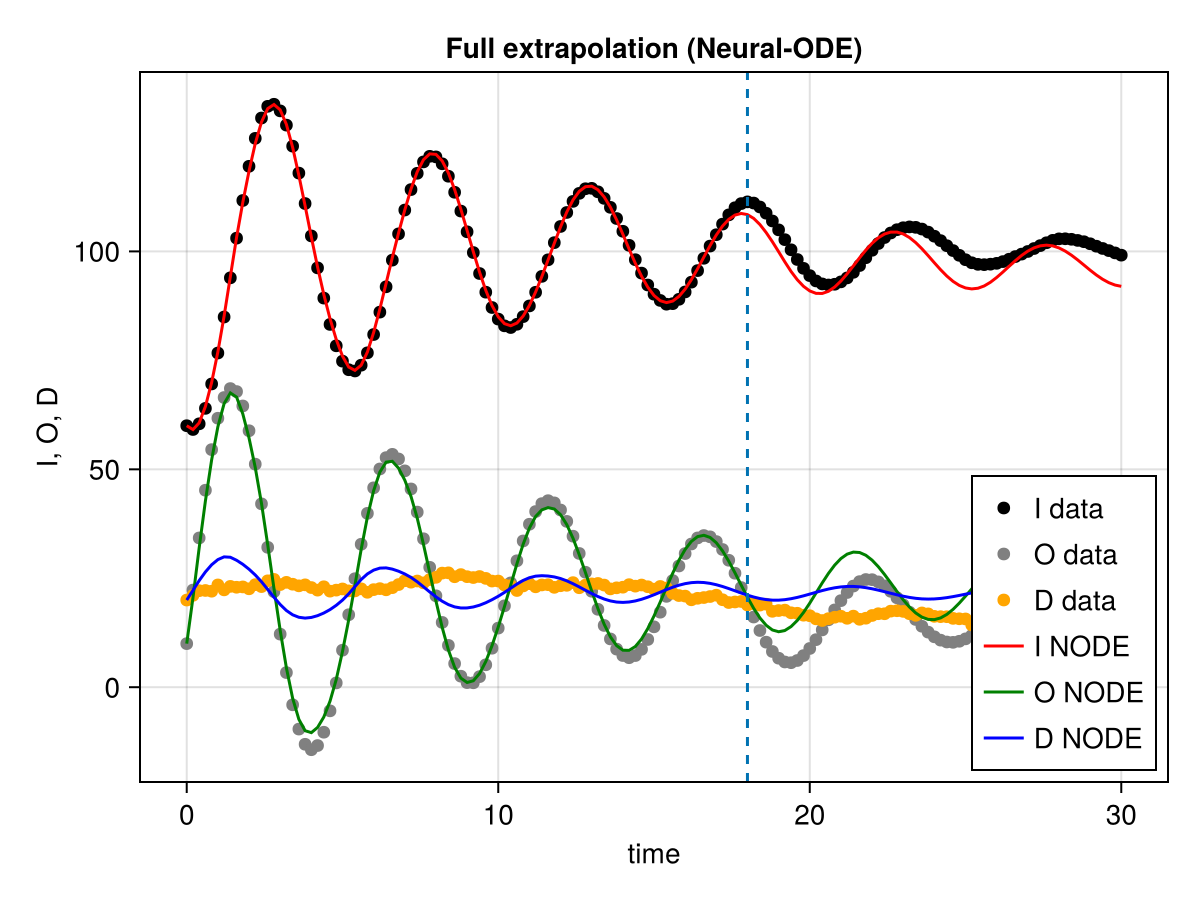}
    \caption{\textbf{NODE forecast.}
    Ground truth (points) and Neural ODE forecasts (lines) for inventory $I$, order rate $O$, and demand $D$ over the full horizon. The dashed vertical line marks the end of the training window ($t\!=\!18$ for a $30$-unit horizon). In the extrapolation region, NODE (full learned RHS) exhibits amplitude/phase drift relative to the true trajectories.}
    \label{fig:ar1-node-forecast-60}
    \end{subfigure}\hfill
    \begin{subfigure}[t]{0.48\textwidth}
    \centering
    \includegraphics[width=\linewidth]{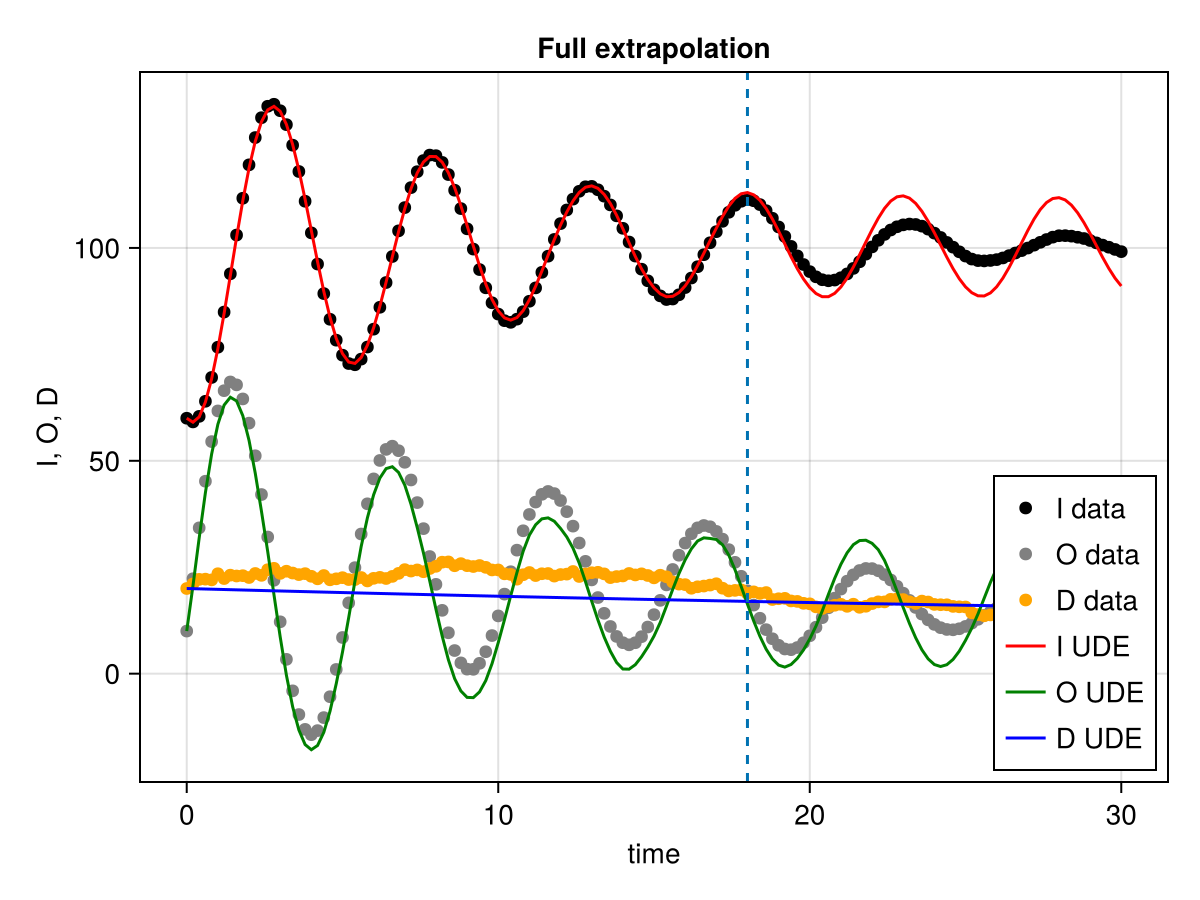}
    \caption{\textbf{UDE forecast.}
    Ground truth (points) and UDE forecasts (lines) for $I$, $O$, and $D$. The dashed vertical line marks the end of training ($t\!=\!18$). The physics-informed UDE preserves the oscillation amplitude and phase more accurately in the forecast window.}
    \label{fig:ar1-ude-forecast-60}
    \end{subfigure}\hfill
\caption{AR(1) demand, 60\% training split}
\label{fig:ar1-forecast-60}
\end{figure}

\begin{figure}[htbp]
\centering
    \begin{subfigure}[t]{0.48\textwidth}
    \centering
    \includegraphics[width=\linewidth]{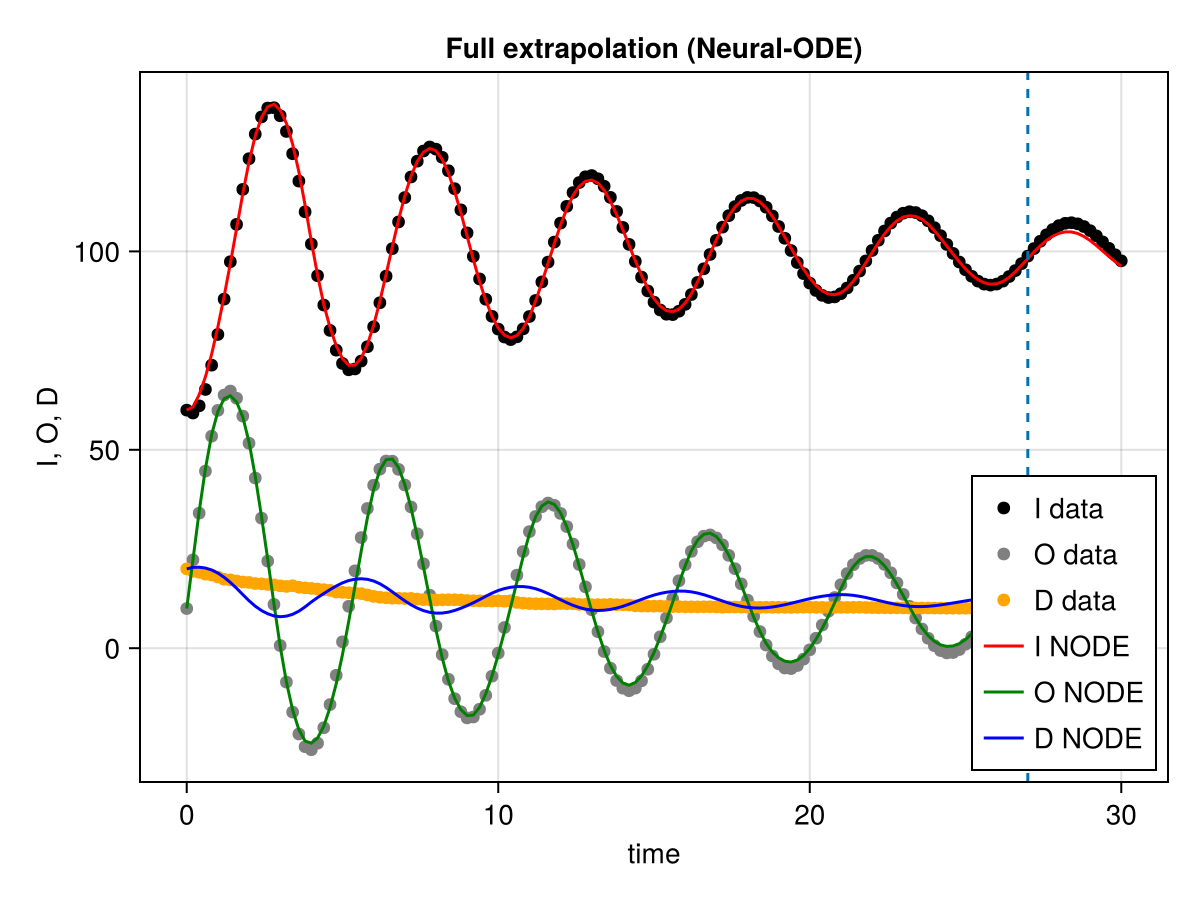}
    \caption{\textbf{NODE forecast.}
    Ground truth (points) and Neural ODE forecasts (lines) for $I$, $O$, and $D$ over the full horizon. The dashed vertical line marks the end of the training window ($t\!=\!27$). For lognormal (heavy-tailed) demand, NODE (full learned RHS) reproduces sharp order oscillations and follows the inventory peak in the extrapolation region.}
    \label{fig:logn-node-forecast-90}
    \end{subfigure}\hfill
    \begin{subfigure}[t]{0.48\textwidth}
    \centering
    \includegraphics[width=\linewidth]{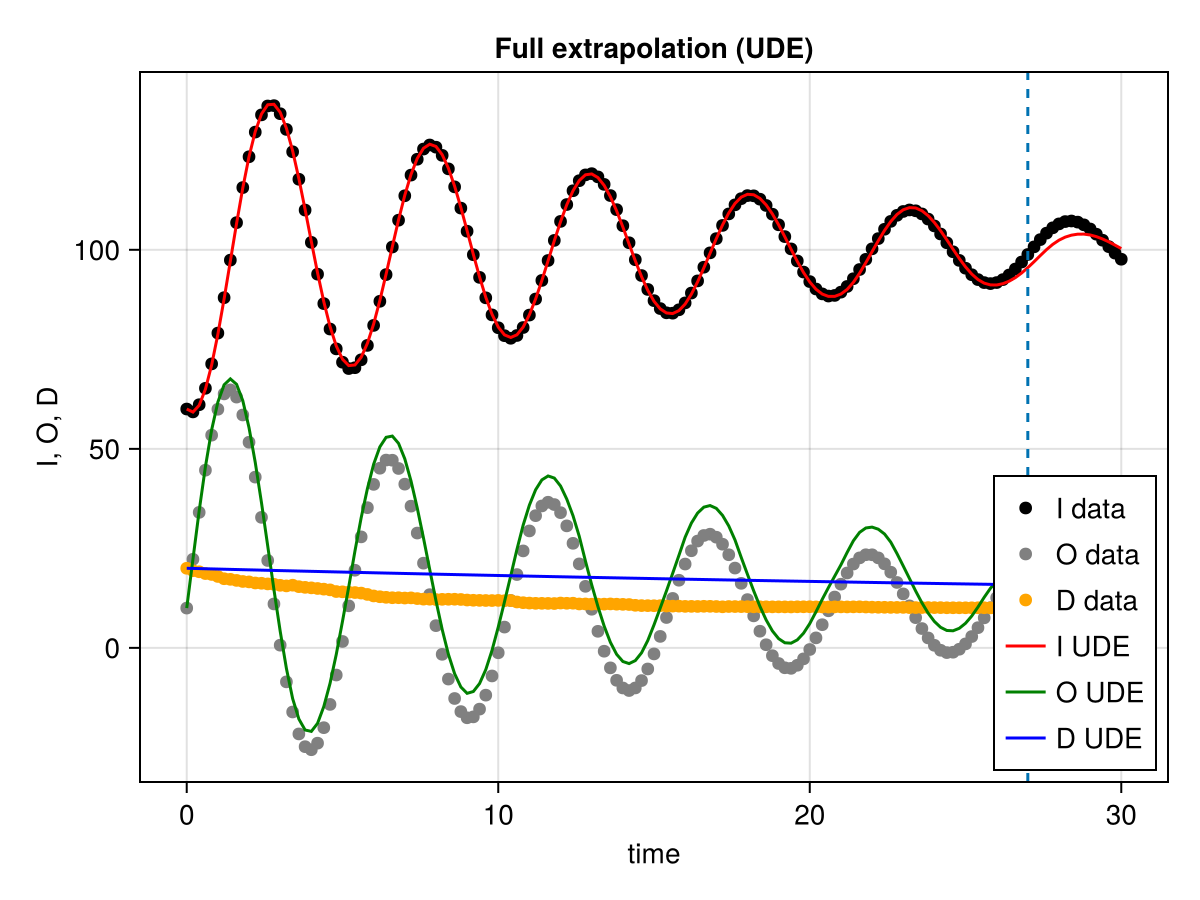}
    \caption{\textbf{UDE forecast.}
    Ground truth (points) and UDE forecast (line) for inventory $I$; the dashed line marks $t\!=\!27$. Under lognormal shocks, the physics-informed UDE remains smoother and slightly underestimates peaks, showing structural bias when rare spikes dominate.}
    \label{fig:logn-ude-forecast-90}
    \end{subfigure}\hfill
\caption{\textbf{Lognormal demand, 90\% training split.} NODE is better at capturing heavy-tailed spikes in the forecast window, whereas UDE tends to under-react due to its small additive correction on $\dot O$}
\label{fig:logn-forecast-90}
\end{figure}

\newpage

\section{Discussion and Conclusion}
\label{sec:conclusion}
Our results clarify when hybrid, structure-preserving models are preferable to fully learned dynamics in modeling inventory dynamics. In regimes with autocorrelation or light-tailed noise, the UDE’s built-in bias - inventory conservation combined with an order-up-to structure - resulted in a lower forecast error and more stable extrapolation than a black-box NODE. This aligns with control-theoretic accounts of bullwhip, where amplification is largely governed by policy structure and lead-time dynamics, not just data volume. In contrast, under heavy-tailed lognormal shocks, the unconstrained NODE captured rare surges more faithfully, indicating that strong structural bias can underreact to extremes when the residual capacity is too limited.

Relative to prior supply-chain work, our results build on classical analyses that quantify variance amplification under order-up-to policies and alternative forecasting rules. Those studies derive closed-form conditions for amplification and show how accumulated lead times and forecasting choices drive the bullwhip effect. We reach similar qualitative conclusions, but in a learning-based context. When the dynamics align with known structure and the noise is moderate, constraints improve stability; when shocks are rare but large, flexibility becomes more important. Compared with recent physics-informed approaches (PINNs, UDEs) and continuous-depth neural models (NODEs), our contribution is to make the trade-off explicit across demand regimes and data budgets within a single controlled testbed.

The study has limits. We simulate a single-echelon setting with fully observed, exogenous demand and omit transport/production delays, capacity constraints, and costs. The residual of the UDE is additive and invariant over time, which can blunt the response to multiplicative effects or regime changes. Our evaluation emphasizes RMSE on states; it does not assess downstream operational metrics (such as service levels, backorders) or decision quality when models close the loop.

Several directions emerge from this work. On the methodological side, richer residual formulations, such as multiplicative or gated variants, along with SDE-UDE hybrids for discontinuous environments and models that capture delayed or distributed-delay dynamics, could help reduce bias under extreme conditions. From a computational perspective, training schemes that combine forecasting with decision making, or objectives that incorporate uncertainty through risk-sensitive or quantile-based losses, may better align learning with operational costs. On the domain side, extending the framework to multi-echelon supply chains with partial observation, examining when information sharing offsets model bias, and validating the approach on real demand traces would provide stronger tests of robustness and practical relevance.

We compared a black-box NODE with a structure-preserving UDE for continuous-time inventory forecasting under stochastic demand. The key result is regime-dependent: with AR(1) or Gaussian noise, enforcing conservation and an order-up-to backbone improves forecast accuracy and stability; with heavy-tailed lognormal shocks, a flexible NODE better tracks spikes. This pattern persisted as training data shrank.

The broader implication is practical guidance on when to impose structural bias: use structure when governing laws are credible and noise is modest; relax structure as tail heaviness and regime variability increase. Beyond inventory dynamics, the same principle applies to other systems where conservation, stability, and rare events coexist, suggesting hybrid models that adapt their level of structural commitment to the noise environment and the task objective.

\end{document}